\documentclass{article}

\usepackage[preprint]{spconf}
\usepackage{amsmath,graphicx}

\usepackage{algorithm}
\usepackage{algorithmic}
\usepackage{rotating}
\usepackage{multirow}
\usepackage{amssymb}
\usepackage{float}
\usepackage{textpos}

\title{ECAP: Extensive Cut-and-Paste Augmentation for Unsupervised Domain Adaptive Semantic Segmentation}

\name{Erik Brorsson$^{\star \dagger}$ \qquad Knut Åkesson$^{\dagger}$ \qquad Lennart Svensson$^{\dagger}$ \qquad Kristofer Bengtsson$^{\star}$}
  
  \address{$^{\star}$ Global Trucks Operations, Volvo Group, Gothenburg, Sweden \\
      $^{\dagger}$Department of Electrical Engineering, Chalmers University of Technology, Gothenburg, Sweden}

\begin{document}



\maketitle


\begin{textblock*}{150mm}(.05\textwidth,-7.5cm)
\noindent
© 2024 IEEE. Personal use of this material is permitted. Permission from IEEE must be obtained for all other uses, in any current or future media, including reprinting/republishing this material for advertising or promotional purposes, creating new collective works, for resale or redistribution to servers or lists, or reuse of any copyrighted component of this work in other works.
\end{textblock*}

\begin{abstract}
We consider unsupervised domain adaptation (UDA) for semantic segmentation in which the model is trained on a labeled source dataset and adapted to an unlabeled target dataset. Unfortunately, current self-training methods are susceptible to misclassified pseudo-labels resulting from erroneous predictions. Since certain classes are typically associated with less reliable predictions in UDA, reducing the impact of such pseudo-labels without skewing the training towards some classes is notoriously difficult. 
To this end, we propose an extensive cut-and-paste strategy (ECAP) to leverage reliable pseudo-labels through data augmentation.
Specifically, ECAP maintains a memory bank of pseudo-labeled target samples throughout training and cut-and-pastes the most confident ones onto the current training batch. We implement ECAP on top of the recent method MIC and boost its performance on two synthetic-to-real domain adaptation benchmarks. Notably, MIC+ECAP reaches an unprecedented performance of 69.1 mIoU on the Synthia$\rightarrow$Cityscapes benchmark. Our code is available at https://github.com/ErikBrorsson/ECAP.
\end{abstract}


\section{Introduction}
\label{sec:intro}
 
    Unsupervised domain adaptation (UDA) is one of many approaches used for semantic segmentation that aims to relax the requirements on the availability of annotated training data.
    In UDA, an unlabeled \textit{target} dataset sampled from the same distribution as the test dataset is available along with a labeled \textit{source} dataset sampled from another distribution. The source dataset could constitute an already annotated dataset or a synthetic dataset for which annotations can easily be created. Due to the difference in distribution between source and target data, also known as the domain gap, a network trained on source data typically does not perform well on target data. Therefore, UDA methods use the unlabeled target data to adapt the model to the test data distribution, for example, through adversarial training \cite{cycada, adaptsegnet, advent, discriminative_path_repr} or self-training \cite{CBST, dacs, iast, proda, hrda, mic}.
    
    In recent years, self-training has dominated the field and is adopted by many recent works \cite{daformer, hrda, chen2022pipa, mic}. 
    A pivotal component of this framework is the DACS \cite{dacs} data augmentation, which entails mixing a source and target image through a cut-and-paste operation. Although DACS augmentation is effective in bridging the domain gap, it does not handle the noise inherent to the pseudo-labels on which the model is trained. Most existing methods that address this problem attempt filtering the pseudo-labels based on predicted confidence scores \cite{CBST, iast, crst, pycda}. Unsurprisingly, the unconfident pseudo-labels often belong to \textit{hard-to-adapt} classes, meaning that such approaches may impede the learning of these classes since the focus is shifted to \textit{easy-to-adapt} classes. Multiple UDA methods facilitate learning hard-to-adapt classes by setting class-wise confidence thresholds when generating/filtering pseudo-labels \cite{CBST, iast, sac} or using sampling schemes that favors these classes \cite{daformer, sac, dsp}. 
    Nevertheless, maintaining a healthy balance between classes while focusing training on reliable pseudo-labels remains a challenge. 
    
    In this work, we take an entirely different approach to dealing with pseudo-label noise in self-training, which is based on cut-and-paste data augmentation. 
    Specifically, we build a memory bank of pseudo-labeled target samples during training, and in each iteration, cut-and-paste confident samples from the memory bank onto the current training batch. By cut-and-pasting content from a large pool of images, our proposed method effectively makes use of the, typically scarce, confident pseudo-labels of the hard-to-adapt classes and shifts focus away from erroneous pseudo-labels during training. Our method, which we call ECAP: Extensive Cut-and-Paste, is to the best of our knowledge the first UDA method for semantic segmentation that 
    aims to increase the proportion of correct pseudo-labels in each training image by means of cut-and-paste augmentation. Through comprehensive evaluation, ECAP is shown to increase the performance of multiple UDA methods based on self-training on the synthetic-to-real domain adaptation task. Notably, we reach new state-of-the-art performance on both the GTA$\rightarrow$Cityscapes and Synthia$\rightarrow$Cityscapes benchmarks by boosting the performance of the recent method MIC \cite{mic} by $0.3$ and $1.8$ mIoU respectively.
    

    Our main contributions are:
    \begin{enumerate}
        \item We propose a data augmentation method for unsupervised domain adaptive semantic segmentation that is designed to counteract pseudo-label noise during self-training
        \item We demonstrate that our approach increases the mIoU score of previous state-of-the-art on two synthetic-to-real domain adaptation benchmarks
        \item We analyze the adverse effect of pseudo-label noise during self-training 
    \end{enumerate}

\section{Related Work}



    \textbf{Unsupervised Domain Adaptation (UDA)}
        methods for semantic segmentation can broadly be categorized as either based on adversarial learning \cite{cycada, adaptsegnet, advent, discriminative_path_repr} or self-training \cite{CBST, iast, dacs, proda, hrda, mic}. In the former category, a discriminator network enforces domain invariance in the input through style-transferred images \cite{cycada, li2019bidirectional} or in the feature/output space of the segmentation network \cite{cycada, li2019bidirectional, adaptsegnet, advent, saito2018maximum}.
        In self-training, on the other hand, pseudo-labels are created for the unlabeled target data on which the model is then trained further. Multiple works also use an adaptation curriculum \cite{cda, pycda}, entropy minimization \cite{advent,chen2019domain, iast, hiast} and more recently contrastive learning \cite{chen2022pipa, huang2022category}.

        Inspired by consistency regularization \cite{sohn2020fixmatch}, many self-training methods enforce consistency between predictions on strongly augmented images with the pseudo-labels generated on weakly-augmented images. In particular, mixing the content of a source and target image is commonly used as the strong augmentation \cite{dacs, dsp, context-aware-mixup}. Furthermore, while pseudo-labels may be generated naively from the model's predictions \cite{CBST, iast}, many methods attempt increasing the quality of the pseudo-labels by e.g., using a mean teacher \cite{dacs, hrda}
        , label prototypes \cite{proda, cag}, or averaging predictions over multiple stochastic forward passes \cite{context-aware-mixup} or augmentations of the input \cite{sac}.

        The noise inherent to the pseudo-labels is often handled by either filtering the unreliable pseudo-labels by predicted confidence \cite{CBST, iast, crst, pycda}, entropy \cite{context-aware-mixup} or uncertainty \cite{uncertainty_aware_consistency_regularization}, or using a weighted loss function that assigns lower weight to unreliable pseudo-labels by e.g., model confidence \cite{sac}, uncertainty \cite{rectifying_with_uncertainty} or depth estimation \cite{corda}. A problem with reducing the contribution of unreliable pseudo-labels is that the loss may become dominated by easy-to-adapt classes that are more reliable than the hard-to-adapt classes (these are often, although not necessarily, long-tail classes). To this end, many methods use class-wise confidence thresholds when generating/filtering pseudo-labels such that hard-to-adapt classes can be prioritized \cite{CBST, iast, sac}. Others suggest using a focal loss to emphasize difficult samples \cite{sac} or data sampling schemes that favors such classes \cite{daformer, sac, dsp}. Different from all previous methods, ECAP aims to mitigate the issue of noisy pseudo-labels by increasing the proportion of correctly pseudo-labeled pixels in the training images by means of cut-and-paste data augmentation.

        \noindent
        \textbf{Cut-and-Paste Data Augmentation}
        has been used both for image classification \cite{cutmix}, object detection \cite{cut_paste_learn, modeling_context_key_to_OD, ge2022empaste}, semantic segmentation \cite{classmix, dacs, hiast} and instance segmentation \cite{ simple_copy_paste_instance_segmentation, zhao2022xpaste, ge2022empaste} in a variety of settings, including supervised learning \cite{cutmix, modeling_context_key_to_OD, simple_copy_paste_instance_segmentation, zhao2022xpaste}, semi-supervised learning \cite{classmix, simple_copy_paste_instance_segmentation} and weakly-supervised learning \cite{cutmix, ge2022empaste}. Notably, recent works \cite{ge2022empaste} and \cite{zhao2022xpaste} leverage generative models and foreground segmentation models to create large-scale datasets by cut-and-pasting multiple object instances into every training image. In UDA however, cut-and-paste augmentation has mainly been limited to methods that mix a single source and target image as a way of overcoming the domain gap. This idea, first proposed by DACS \cite{dacs} has been used in many succeeding works \cite{daformer, hrda, mic, chen2022pipa} and given rise to a number of variations \cite{dsp, context-aware-mixup, ida}. Unlike all previous methods for UDA, ECAP is not restricted to images in a single batch but rather performs cut-and-paste data augmentation using a large pool of target images that are stored in a memory bank. Importantly, this allows ECAP to select the most suitable images for augmentation and training by considering the confidence of the associated pseudo-label.

\section{Preliminary}
\label{preliminary}

    In this section, we establish the notation and provide relevant details for the 
    self-training framework used in \cite{dacs, daformer, hrda, mic} on top of which ECAP is implemented.
    
         In self-training, a segmentation network $f_{\theta}$ (called the student) is trained on a set of $N_S$ source domain images with associated labels $\{x^{S,k}, y^{S,k}\}_{k=1}^{N_S}$ and on a set of $N_T$ unlabeled target domain images with associated pseudo-labels $\{x^{T,k}, \tilde{y}^{T,k}\}_{k=1}^{N_T}$. Dropping the index $k$, for ease of notation, the pseudo-label $\tilde{y}$ of an image $x^T$ constitutes a one-hot vector $\tilde{y}_{ij}$ of length $C$ at each pixel location $(i,j)$, where $C$ is the number of classes. The pseudo-label is created during training by the teacher network $g_{\phi}$, which is an exponential moving average of the student's weights. Formally, the pseudo-label is defined by
         

        \begin{equation}
        \label{eq:pseudo_label}
            \tilde{y}_{ijc}^T = [ c = \text{argmax}_{c'} g_{\phi}(x^T)_{ijc'} ] ,
        \end{equation}
         where $[\cdot ]$ denotes the Iverson bracket. 

         In every training iteration, a source and target image and their corresponding label and pseudo-label are mixed by a cut-and-paste operation wherein pixels given by a binary mask $m$ are cut from the source sample and pasted onto the target sample. For convenience, we define such a mixing operation between two images $x_1$ and $x_2$ as
            \begin{equation}
        \label{equation:alpha_mixing}
            \alpha(x_1, x_2, m) := m \odot x_1 + (1 - m) \odot x_2,
        \end{equation}
        where $\odot$ denotes element-wise multiplication. Given the operator $\alpha$ and binary mask $m$, the mixed image and corresponding label is constructed by $x^M = \alpha(x^S, x^T, m)$ and $y^M = \alpha(y^S, \tilde{y}^T, m)$.
        The binary mask $m$ is created by randomly selecting half of the classes that are present in $y^S$ and setting $m_{ij}=1$ for all pixels included in the selections and $m_{ij}=0$ otherwise.

        Given the source and mixed image with corresponding labels as defined above, the weights $\theta$ of the student network $f_{\theta}$ are trained to minimize the loss 

        \begin{equation}
         L(\theta) = \mathbb{E}[\mathcal{L}^S(y^{S}, f_{\theta}(x^S)) + \mathcal{L}^T(y^M, f_{\theta}(x^M), q^M )] ,
        \label{DACS_loss}
        \end{equation}
        where $\mathcal{L}^S$ is the standard cross-entropy loss, $\mathcal{L}^T$ is a weighted cross-entropy loss, and the expectation is taken over data from the source and target datasets. Specifically, 
        the second term is given by

        \begin{equation}
            \mathcal{L}^T(y^M, \hat{y}^M, q^M ) = - \sum_{i=1}^{H} \sum_{j=1}^{W} q^M_{ij} \sum_{c=1}^C y_{ijc}^M log(\hat{y}^M_{ijc} ),
        \label{eq:mixed_loss}
        \end{equation}
        where $W$ and $H$ are the width and height of the image $x^M$ respectively, $\hat{y}^M$ denotes the predictions $f_{\theta}(x^M)$ of the student network, and $q_{ij}^M$ is the weight associated with each pixel. In \cite{dacs,daformer, hrda, mic}, $q_{ij}^M$ equals $1.0$ for pixels originating from the source domain and otherwise equals the ratio of pixels in the target image for which the confidence of the pseudo-label exceeds a threshold $\tau$, i.e.,

        \begin{equation}
            q_{ij}^M = 
            \begin{cases}
1, & \text{if } m_{ij}=1 \\
\frac{\sum_{i=1}^{H} \sum_{j=1}^{W} [(max_{c'} g_{\phi}(x^T)_{ijc'}) > \tau]}{H\cdot W}, & \text{otherwise.}
        \end{cases}
        \label{eq:qM}
        \end{equation}

\section{Extensive Cut-and-Paste (ECAP)}

\begin{figure*}
        \begin{center}
        \includegraphics[scale=0.75]{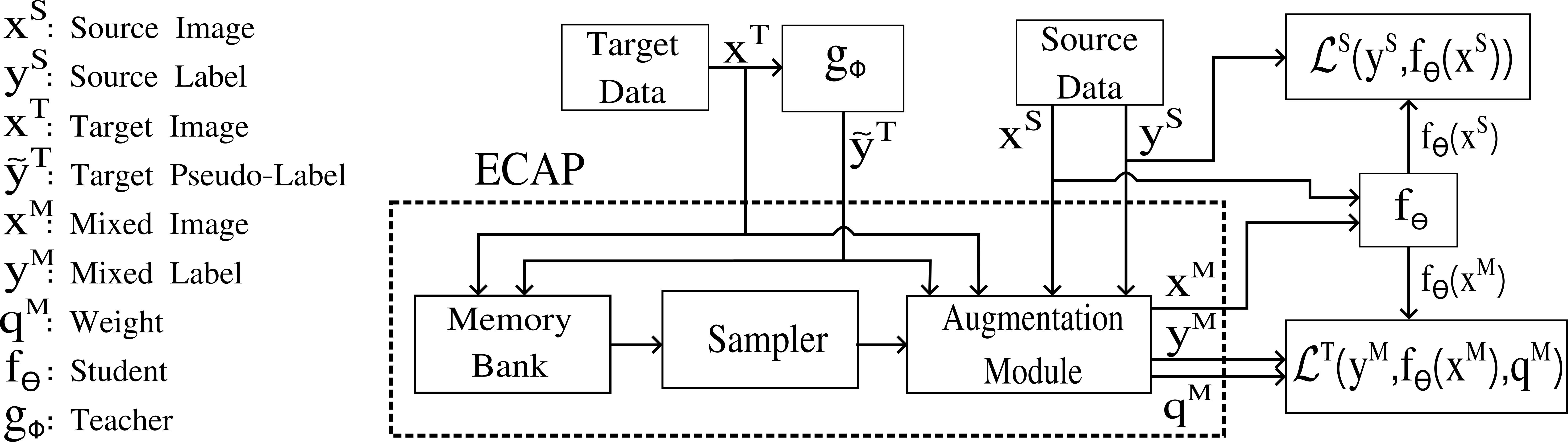}
        \end{center}
           \caption{Schematic illustration of ECAP, constituting a memory bank, a sampler, and an augmentation module, integrated with the self-training framework. The input to ECAP is a source and target image, $x^S$ and $x^T$, along with the corresponding label $y^S$ and pseudo-label $\tilde{y}^T$, which is produced by the teacher $g_{\Phi}$. This input is processed by the augmentation module together with samples from the memory bank to generate a mixed image $x^M$, an associated label $y^M$, and a weight $q^M$. Simultaneously, $x^T$ and $\tilde{y}^T$ are added to the memory bank for future use. The student network $f_{\theta}$ processes both the source image $x^S$ and the mixed image $x^M$ and is supervised by the loss $\mathcal{L}^S(y^{S}, f_{\theta}(x^S)) + \mathcal{L}^T(y^M, f_{\theta}(x^M), q^M )$ during training.}
        \label{fig:ecap-detailed}
        \end{figure*}

    In this work, we propose to 
    cut-and-paste confident pseudo-labeled content from multiple target domain images into the current training batch. Our method ECAP comprises three main components that are described in the following sections: (1) a memory bank containing pseudo-labeled target samples, (2) a sampler drawing samples associated with high confidence from the memory bank, and (3) an augmentation module that creates the augmented training images. We integrate ECAP with the self-training framework detailed in Section \ref{preliminary}, but also recognize that ECAP may be applied in virtually any UDA method based on self-training. Figure \ref{fig:ecap-detailed} provides a schematic illustration of our method.


        \subsection{Memory Bank}
            \label{implementation:memory_bank}
            During training, a memory bank $B_c$ is constructed for each class $c$ in the dataset.
            Each memory bank $B_c$ consists of a set of $\vert B_c \vert$ images, pseudo-labels and confidence scores $\{(x^{B_c, l},y^{B_c, l}, q^{B_c, l}) \}_{l=1}^{\vert B_c \vert}$. Again, dropping the index $l$ for ease of notation, $x^{B_c}$ and $y^{B_c}$ are constructed by applying a binary mask $m_c$ to a target image $x^T$ and pseudo-label $\tilde{y}^T$ according to $( x^{B_c}, y^{B_c} ) = (x^{T} \odot m_c, \tilde{y}^{T}  \odot m_c)$. The binary mask $m_c$ takes the value $1.0$ at each pixel location $(i,j)$ for which $\tilde{y}^{T}_{ij}$ is of class $c$, and otherwise $m_c$ takes the value $0$. Furthermore, the confidence score $q^{B_c}$ is defined as the average confidence of the teacher's predictions corresponding to pseudo-label of class $c$ for the target image $x^T$ according to

            \begin{equation}
                q_{}^{B_c} = \frac{ \sum_{i=1}^{W} \sum_{j=1}^{H} g_{\phi}(x^T)_{ijc} [c = \text{argmax}_{c'}(g_{\phi}(x^T)_{ijc'}) ] } {\sum_{i=1}^{W} \sum_{j=1}^{H} [c = \text{argmax}_{c'}(g_{\phi}(x^T)_{ijc'}) ]} .
            \label{quality_i}
            \end{equation}
            Note that $x^{B_c},y^{B_c}$ and $q^{B_c}$ are computed for every class $c$ present in $\tilde{y}^T$, which implies that (different parts of) a single target image $x^T$ and the corresponding pseudo-label $\tilde{y}^T$ are often added to multiple memory banks. 
            To avoid storing duplicates of certain target samples as training spans multiple epochs, we allow at most one sample in each memory bank $B_c$ per target image.

        \subsection{Sampler}
        \label{method:sampling_scheme}
            In each training iteration, a set of samples  $\{S^{B_c, l}\}_{c,l} = \\ \{(x^{B_c, l}, y^{B_c, l})\}_{c,l}$ is obtained by
            first selecting a set of classes and then drawing a sample from the memory bank of each selected class (we use a slight abuse of notation here as the index pair $c, l$ takes on values that depend on the outcome of the sampling). Formally, we define a set of random variables $\{\mathbf{r_c} \sim \text{Bern}(p_{r_c}) \}_{c=1}^C$ of which a set of observations $\{r_c\}_{c=1}^C$ is obtained in each training iteration. For every class $c$, one sample is then drawn from memory bank $B_c$ if $r_c = 1$. Since samples of high quality are preferred, we use the confidence value $q^{B_c}$, defined in Equation \ref{quality_i}, when sampling from each memory bank. Specifically, we propose uniform sampling from the top $n^{B_c}$ most confident samples in each memory bank $B_c$, where $n^{B_c}$ is a hyperparameter. Consequently, if we sort the confidence scores of $B_c$ in descending order and let $q(n^{B_c})$ be the score at position $n^{B_c}$, the sampling probability $p^{B_c, l}$ of each sample in $B_c$ can be defined as 
            \begin{equation}
                p^{B_c, l}_{} =
            \begin{cases}
1 / n^{B_c}, & \text{if } q^{B_c,l} \geq q(n^{B_c})\\
0, & \text{otherwise.}
            \end{cases}
            \label{eq:p-bc-i}
            \end{equation}
            
            Furthermore, we argue that the sampling probability $p_{r_c}$ should increase as training progresses and the quality of the memory bank increases. In this work, we let $p_{r_c}$ be equal across all classes and define $p_{r_c} = n_0 \sigma(\frac{MEC-\beta}{\gamma})$, where $n_0 \in [0,1]$ is a constant, $MEC$ is the mean expected confidence of a sample drawn uniformly from the set of memory banks $\{B\}_c$, $\sigma$ is the \textit{sigmoid} function, and $\beta$ and $\gamma$ are two hyperparameters. With these design choices, $p_{r_c}$ is given by
            \begin{equation}
            \label{eq:prc_final}
                p_{r_c} = n_0 \sigma (\frac{ \frac{1}{C}\sum_{c=1}^C \sum_{l=1}^{\vert B_c \vert} q^{B_c, l}_{} p^{B_c, l}_{} - \beta}{\gamma}),
            \end{equation}
            where the index $l$ runs over every sample (with associated confidence value $q^{B_c, l}$ and sampling probability $p^{B_c, l}$) in memory bank $B_c$. Note that $\beta$ can be thought of as a typical value of $MEC$ at which ECAP sampling comes online, while $\gamma$ relates to how quickly sampling comes online as $MEC$ approaches $\beta$. 

        \subsection{Augmentation Module}

            The set of samples $\{S^{B_c, l}\}_{c,l}$ obtained by the sampler in Section \ref{method:sampling_scheme} is used to create a composite image $x^B$ and corresponding pseudo-label $y^B$. This is done by initializing $x^B$ and $y^B$ as blank canvases, iterating over the samples in $\{S^{B_c, l}\}_{c,l}$, and in each iteration cut-and-pasting the content of $S^{B_c, l}$ corresponding to class $c$ onto the canvases. Before pasting, however, the sample is subject to a series of transformations, including random scaling, translation and horizontal flipping. The construction of $x^B$ and $y^B$ is described in Algorithm \ref{alg:composite-image}, where $\alpha$ is the mixing operator defined in Equation \ref{equation:alpha_mixing}.

            \begin{algorithm}
            \caption{Creating composite image}
            \label{alg:composite-image}
            \begin{algorithmic}
            \STATE Initialize $x^B$ and $y^B$ as blank canvases.
            \STATE $\{S^{B_c, l}\}_{c, l} \gets $ a set of samples provided by the Sampler.
            \STATE Shuffle $\{S^{B_c, l}\}_{c, l}$.
            \FOR{$(x^{B_c, l}, y^{B_c, l})$ in $\{S^{B_c, l}\}_{c,l}$}
                \STATE $T(x^{B_c, l}), T(y^{B_c, l}) \gets$ Applying a set of random transformations to the sample.
                \STATE $m_c \gets T(y^{B_c, l}) = c$
                \STATE $x^B \gets \alpha(T(x^{B_c, l}), x^B, m_c)$
                \STATE $y^B \gets \alpha(T(y^{B_c, l}), y^B, m_c)$
            \ENDFOR
            \RETURN $x^{B}$, $y^B$
            \end{algorithmic}
            \end{algorithm}
            The composite image $x^B$ and corresponding pseudo-label $y^B$ are then used to alter the mixed image $x^M$ and corresponding pseudo-label $y^M$, which are used in self-training as detailed in Section \ref{preliminary}. Specifically, we paste $x^B$ (and $y^B$) onto the source image $x^S$ (and label $y^S$) \textit{prior} to DACS mixing. 
            Formally, if we let $\text{DACS}(x^S, x^T, y^S, \tilde{y}^T)$ denote the algorithm detailed in Section \ref{preliminary} to generate a mixed sample and associated weight from a source and target sample, we let 
            \begin{equation}
            \label{eq:xm_pm}
            \begin{aligned}                
                & (x^M, y^M, q^M) = \\ & \text{DACS}(\alpha (x^B, x^S, m^B), x^T, \alpha (y^B, y^S, m^B), \tilde{y}^T),
            \end{aligned}
            \end{equation}
            where again $\alpha$ denotes the mixing operator defined in Equation \ref{equation:alpha_mixing} and $m^B$ is a binary mask corresponding to the parts of $x^B$ that has been populated. Note that $m^B$ can be trivially computed by a small addition to Algorithm \ref{alg:composite-image}, although we exclude it for simplicity. Also note that the binary mask used internally in the DACS algorithm to construct $x^M$, $y^M$, and $q^M$ is now derived by selecting half of the classes in $ \alpha (y^B, y^S, m^B)$, which means that pixels in $x^M$ originating from the composite image $x^B$ will be associated with a weight of $1.0$. This design choice is made since the pseudo-label of $x^B$ is expected to be of high quality.

\section{Experiments}

    \subsection{Implementation Details}
    \label{exp:implementation}
    \textbf{Datasets:} We study ECAP in the setting of synthetic-to-real domain adaptation on the popular benchmarks GTA$\rightarrow$Cityscapes and Synthia$\rightarrow$Cityscapes. 
    The GTA \cite{Richter_2016_ECCV} and Synthia \cite{Ros_2016_CVPR} datasets consist of 24,966 and 9,000 simulated training images respectively along with accompanying labels. The Cityscapes \cite{cityscapes} training dataset consists of 2,975 images, which are used as the unlabeled target dataset. In line with previous works, we evaluate the performance on the Cityscapes validation set of 500 images. Additionally, we study day-to-nighttime and clear-to-adverse-weather domain adaptation on the benchmarks Cityscapes$\rightarrow$DarkZurich and Cityscapes$\rightarrow$ACDC respectively. The DarkZurich \cite{darkzurich} dataset is captured during nighttime and consist of 2,416 training and 151 test images, while the ACDC \cite{acdc} dataset is captured in adverse weather and comprises 1,600 training and 2,000 test images. 
    The Cityscapes training dataset is used as the labeled source dataset for these benchmarks.
    
    \noindent
    \textbf{Training:} In this work, MIC \cite{mic} is used as a baseline on top of which ECAP is implemented. MIC is based on the self-training framework detailed in Section \ref{preliminary} with the addition of Rare Class Sampling \cite{daformer}, ImageNet Feature Distance \cite{daformer} and multi-resolution fusion \cite{hrda}, as well as Masked Image Consistency \cite{mic}. We use the exact training parameters of MIC \cite{mic} and refer the reader to \cite{mic} for details. The only difference when applying ECAP is that the mixed training sample is created according to Equation \ref{eq:xm_pm}. 
    
    For simplicity, we let $n^{B_c}$ in Equation \ref{eq:p-bc-i} be equal for all classes and set $\gamma = 0.005$ in Equation \ref{eq:prc_final} in all our experiments. In the augmentation module of ECAP, we apply random scaling by a factor $r_s$ sampled uniformly on the interval $[0.1, 1.0]$, as well as random translation and horizontal flipping. In the experiments on Synthia, we disable sampling from the ECAP memory banks corresponding to the classes \textit{Terrain}, \textit{Truck} and \textit{Train} since these don't exist in the Synthia dataset.
    Furthermore, while MIC \cite{mic} refrains from training on pseudo-labels in the region of the Cityscapes images corresponding to the ego-vehicle hood as well as on the image borders, we find this detrimental in the case of Synthia$\rightarrow$Cityscapes. Therefore, we disable this feature and instead train on the entire Cityscapes image for this benchmark. This is elaborated further in the supplementary material available at https://sigport.org/documents/ecap-supplementary.

    \subsection{Comparison with State-of-the-Art}
    \label{exp:sota}
    
        \tabcolsep=0.11cm
        \begin{table*}[h]

        \caption{Semantic segmentation performance (IoU in \%) on four different UDA benchmarks.}
        \label{result:tab1}
        \resizebox{\linewidth}{!}{
        \begin{tabular}{lllllllllllllllllllll}
        \hline
        \multicolumn{1}{l|}{Method}    & \begin{sideways} Road \end{sideways} & \begin{sideways}Sidew.\end{sideways}  & \begin{sideways}Build.\end{sideways} & \begin{sideways}Wall\end{sideways} & \begin{sideways}Fence\end{sideways} & \begin{sideways}Pole\end{sideways} & \begin{sideways}Tr.Light\end{sideways} & \begin{sideways}Tr.Sign\end{sideways} & \begin{sideways}Veget.\end{sideways} & \begin{sideways}Terrain\end{sideways} & \begin{sideways}Sky\end{sideways}  & \begin{sideways}Person\end{sideways} & \begin{sideways}Rider\end{sideways} & \begin{sideways}Car\end{sideways}  & \begin{sideways}Truck\end{sideways} & \begin{sideways}Bus\end{sideways}  & \begin{sideways}Train\end{sideways} & \begin{sideways}M.bike\end{sideways} & \begin{sideways}Bike\end{sideways}  & \multicolumn{1}{|l}{mIoU}     \\ \hline
        \multicolumn{21}{c}{GTA$\rightarrow$Cityscapes (Val.)}                                                                                                                                                                                                                                                                                                                                                                                                                                                                                                                                                                                                                                                                                                                                                                                                                                                                                                                                                                                                                                                                                                                                                                                                                                                                                                     \\ \hline
        \multicolumn{1}{l|}{DACS \cite{dacs}}      & 89.9                                                         & 39.7                                                         & 87.9                                                         & 30.7                                                         & 39.5                                                        & 38.5                                                         & 46.4                                                         & 52.8                                                         & 88.0                                                         & 44.0                                                         & 88.8                                                & 67.2                                                         & 35.8                                                         & 84.5                                                         & 45.7                                                         & 50.2                                                         & 0.0                                                  & 27.3                                                         & \multicolumn{1}{l|}{34.0}                                                         & 52.1                                                               \\

        \multicolumn{1}{l|}{HRDA \cite{hrda}}      & 96.4                                                         & 74.4                                                         & 91.0                                                         & 61.6                                                & 51.5                                                        & 57.1                                                         & 63.9                                                         & 69.3                                                         & 91.3                                                         & 48.4                                                         & 94.2                                                & 79.0                                                         & 52.9                                                         & 93.9                                                         & 84.1                                                         & 85.7                                                         & 75.9                                                 & 63.9                                                         & \multicolumn{1}{l|}{67.5}                                                         & 73.8                                                               \\

            \multicolumn{1}{l|}{PiPa \cite{chen2022pipa}}      & 96.8                                                & 76.3                                                & 91.6                                                & \textbf{63.0}                                                  & 57.7                                                & 60.0                                                  & 65.4                                                 & \textbf{72.6}                                                & \textbf{91.7}                                                & 51.8                                                & \textbf{94.8}                                                 & 79.7                                                & 56.4                                                & 94.4                                                & \textbf{85.9}                                                & 88.4                                                & 78.9                                                & 63.5                                                & \multicolumn{1}{l|}{67.2}                                                                     & 75.6                                                \\

        \multicolumn{1}{l|}{MIC \cite{mic}}       & \textbf{97.4}                                                & 80.1                                                & \textbf{91.7}                                                & 61.2                                                         & 56.9                                              & 59.7                                                         & \textbf{66.0}                                                           & 71.3                                                         & \textbf{91.7}                                                         & 51.4                                                         & 94.3                                       & 79.8                                                         & 56.1                                                         & \textbf{94.6}                                                & 85.4                                                         & 90.3                                                         & 80.4                                        & 64.5                                                         & \multicolumn{1}{l|}{\textbf{68.5}}                                                & 75.9                                                      \\
        
        \multicolumn{1}{l|}{MIC+ECAP}   & \textbf{\begin{tabular}[c]{@{}l@{}}97.4\\ ±0.1\end{tabular}} & \textbf{\begin{tabular}[c]{@{}l@{}}80.3\\ ±0.4\end{tabular}} & \begin{tabular}[c]{@{}l@{}}91.6\\ ±0.0\end{tabular} & \begin{tabular}[c]{@{}l@{}}60.4\\ ±1.9\end{tabular} & \textbf{\begin{tabular}[c]{@{}l@{}}58.2\\ ±0.6\end{tabular}} & \textbf{\begin{tabular}[c]{@{}l@{}}60.9\\ ±0.7\end{tabular}} & \begin{tabular}[c]{@{}l@{}}65.6\\ ±0.6\end{tabular} & \begin{tabular}[c]{@{}l@{}}71.8\\ ±1.0\end{tabular} & \textbf{\begin{tabular}[c]{@{}l@{}}91.7\\ ±0.1\end{tabular}} & \textbf{\begin{tabular}[c]{@{}l@{}}52.8\\ ±0.5\end{tabular}} & \begin{tabular}[c]{@{}l@{}}93.9\\ ±0.2\end{tabular} & \textbf{\begin{tabular}[c]{@{}l@{}}80.6\\ ±0.5\end{tabular}} & \textbf{\begin{tabular}[c]{@{}l@{}}57.2\\ ±1.4\end{tabular}} & \begin{tabular}[c]{@{}l@{}}94.4\\ ±0.1\end{tabular} & \begin{tabular}[c]{@{}l@{}}85.2\\ ±2.2\end{tabular} & \textbf{\begin{tabular}[c]{@{}l@{}}91.1\\ ±0.1\end{tabular}} & \textbf{\begin{tabular}[c]{@{}l@{}}82.1\\ ±1.4\end{tabular}} & \textbf{\begin{tabular}[c]{@{}l@{}}65.2\\ ±0.4\end{tabular}} & \multicolumn{1}{l|}{\begin{tabular}[c]{@{}l@{}}67.8\\ ±0.7\end{tabular}} & \textbf{\begin{tabular}[c]{@{}l@{}}76.2\\ ±0.1\end{tabular}} \\ \hline
        \multicolumn{21}{c}{Synthia$\rightarrow$Cityscapes (Val.)}                                                                                                                                                                                                                                                                                                                                                                                                                                                                                                                                                                                                                                                                                                                                                                                                                                                                                                                                                                                                                                                                                                                                                                                                                                                                                                 \\ \hline
        \multicolumn{1}{l|}{DACS \cite{dacs}}      & 80.6                                                         & 25.1                                                         & 81.9                                                         & 21.5                                                         & 2.9                                                         & 37.2                                                         & 22.7                                                         & 24.0                                                         & 83.7                                                         &                                                              & 90.8                                                & 67.6                                                         & 38.3                                                         & 82.9                                                         &                                                              & 38.9                                                         &                                                      & 28.5                                                         & \multicolumn{1}{l|}{47.6}                                                         & 48.3                                                               \\
        \multicolumn{1}{l|}{HRDA \cite{hrda}}      & 85.2                                                         & 47.7                                                         & 88.8                                                         & 49.5                                                         & 4.8                                                         & 57.2                                                         & 65.7                                                         & 60.9                                                         & 85.3                                                         &                                                              & 92.9                                                & 79.4                                                         & 52.8                                                         & 89.0                                                           &                                                              & 64.7                                                         &                                                      & 63.9                                                         & \multicolumn{1}{l|}{64.9}                                                         & 65.8                                                               \\

\multicolumn{1}{l|}{PiPa \cite{chen2022pipa}}  & 88.6                                                & 50.1                                                & \textbf{90.0}                                                  & \textbf{53.8}                                                & 7.7                                                 & 58.1                                                & 67.2                                                 & 63.1                                                & 88.5                                                &                                                    & 94.5                                                 & 79.7                                                & 57.6                                                & 90.8                                                &                                                    & 70.2                                                &                                                    & 65.1                                                & \multicolumn{1}{l|}{\textbf{66.9}}                                                                     & 68.2      \\                                         

        \multicolumn{1}{l|}{MIC \cite{mic}}       & 86.6                                                         & 50.5                                                         & 89.3                                                         & 47.9                                                         & 7.8                                                         & \textbf{59.4}                                                         & 66.7                                                         & \textbf{63.4}                                                         & 87.1                                                &                                                              & 94.6                                       & 81.0                                                           & \textbf{58.9}                                                & 90.1                                                         &                                                              & 61.9                                                         &                                                      & \textbf{67.1 }                                                        & \multicolumn{1}{l|}{64.3}                                                         & 67.3                                                               \\
        \multicolumn{1}{l|}{MIC$\dagger$ }  & \textbf{\begin{tabular}[c]{@{}l@{}}91.0\\ ±0.9\end{tabular}} & \textbf{\begin{tabular}[c]{@{}l@{}}55.7\\ ±2.0\end{tabular}} & \begin{tabular}[c]{@{}l@{}}89.9\\ ±0.1\end{tabular} & \begin{tabular}[c]{@{}l@{}}50.4\\ ±1.7\end{tabular} & \textbf{\begin{tabular}[c]{@{}l@{}}8.4\\ ±0.1\end{tabular} } & \begin{tabular}[c]{@{}l@{}}58.8\\ ±0.4\end{tabular} & \begin{tabular}[c]{@{}l@{}}66.7\\ ±0.2\end{tabular} & \begin{tabular}[c]{@{}l@{}}62.9\\ ±0.4\end{tabular} & \textbf{\begin{tabular}[c]{@{}l@{}}89.2\\ ±1.2\end{tabular}} &   & \begin{tabular}[c]{@{}l@{}}94.6\\ ±0.1\end{tabular} & \begin{tabular}[c]{@{}l@{}}81.2\\ ±0.1\end{tabular} & \begin{tabular}[c]{@{}l@{}}57.6\\ ±0.5\end{tabular} & \begin{tabular}[c]{@{}l@{}}90.6\\ ±0.3\end{tabular} &  & \begin{tabular}[c]{@{}l@{}}65.3\\ ±5.7\end{tabular} &   & \begin{tabular}[c]{@{}l@{}}66.0\\ ±0.7\end{tabular} & \multicolumn{1}{l|}{\begin{tabular}[c]{@{}l@{}}63.2\\ ±1.7\end{tabular}} & \begin{tabular}[c]{@{}l@{}}68.2\\ ±0.2\end{tabular} \\
        
        \multicolumn{1}{l|}{MIC$\dagger$+ECAP}  & \begin{tabular}[c]{@{}l@{}}90.8\\ ±0.8\end{tabular} & \begin{tabular}[c]{@{}l@{}}55.2\\ ±1.6\end{tabular} & \textbf{\begin{tabular}[c]{@{}l@{}}90.0\\ ±0.1\end{tabular}} & \begin{tabular}[c]{@{}l@{}}50.7\\ ±3.7\end{tabular} & \begin{tabular}[c]{@{}l@{}}8.2\\ ±0.6\end{tabular}  & \begin{tabular}[c]{@{}l@{}}59.3\\ ±0.2\end{tabular} & \textbf{\begin{tabular}[c]{@{}l@{}}68.3\\ ±0.3\end{tabular}} & \begin{tabular}[c]{@{}l@{}}63.0\\ ±0.2\end{tabular} & \begin{tabular}[c]{@{}l@{}}89.0\\ ±0.3\end{tabular} &   & \textbf{\begin{tabular}[c]{@{}l@{}}94.8\\ ±0.2\end{tabular}} & \textbf{\begin{tabular}[c]{@{}l@{}}81.8\\ ±0.3\end{tabular}} & \begin{tabular}[c]{@{}l@{}}58.6\\ ±0.3\end{tabular} & \textbf{\begin{tabular}[c]{@{}l@{}}90.9\\ ±0.0\end{tabular}} &   & \textbf{\begin{tabular}[c]{@{}l@{}}71.4\\ ±0.4\end{tabular}} &  & \textbf{\begin{tabular}[c]{@{}l@{}}67.1\\ ±1.3\end{tabular}} & \multicolumn{1}{l|}{\begin{tabular}[c]{@{}l@{}}65.9\\ ±0.5\end{tabular}} & \textbf{\begin{tabular}[c]{@{}l@{}}69.1\\ ±0.3\end{tabular}} \\ \hline                              

        \multicolumn{21}{c}{Cityscapes$\rightarrow$Dark Zurich (Test)}                                                                                                                                                                                                                                                                                                                                                  \\ \hline
        \multicolumn{1}{l|}{DAFormer \cite{daformer}}         & 93.5                                                & 65.5                                                & 73.3                                                & 39.4                                                & 19.2                                                & 53.3                                                & 44.1                                                & 44.0                                                & 59.5                                                & 34.5                                                & 66.6                                                & 53.4                                                & 52.7                                                & 82.1                                                & 52.7                                                & 9.5                                                 & 89.3                                                & 50.5                                                & \multicolumn{1}{l|}{38.5}                                                & 53.8                                                \\
        \multicolumn{1}{l|}{HRDA \cite{hrda}}               & 90.4                                                & 56.3                                                & 72.0                                                & 39.5                                                & 19.5                                                & 57.8                                                & \textbf{52.7}                                                & 43.1                                                & 59.3                                                & 29.1                                                & 70.5                                                & 60.0                                                & 58.6                                                & \textbf{84.0}                                                & \textbf{75.5}                                                & 11.2                                                & 90.5                                                & 51.6                                                & \multicolumn{1}{l|}{40.9  }                                                                   & 55.9      \\                                         
        
        \multicolumn{1}{l|}{MIC \cite{mic}}                 & \textbf{94.8}                                                & \textbf{75.0}                                                & \textbf{84.0}                                                & \textbf{55.1}                                                & \textbf{28.4}                                                & 62.0                                                & 35.5                                                & \textbf{52.6}                                                & 59.2                                                & \textbf{46.8}                                                & 70.0                                                & \textbf{65.2}                                                & \textbf{61.7}                                                & 82.1                                                & 64.2                                                & \textbf{18.5}                                                & 91.3                                                & \textbf{52.6}                                                & \multicolumn{1}{l|}{44.0}                                                & 60.2                                                \\
        
        \multicolumn{1}{l|}{MIC+ECAP} & \begin{tabular}[c]{@{}l@{}}90.6\\ ±2.3\end{tabular} & \begin{tabular}[c]{@{}l@{}}55.8\\ ±9.5\end{tabular} & \begin{tabular}[c]{@{}l@{}}82.2\\ ±1.4\end{tabular} & \begin{tabular}[c]{@{}l@{}}53.4\\ ±4.4\end{tabular} & \begin{tabular}[c]{@{}l@{}}25.0\\ ±3.1\end{tabular} & \textbf{\begin{tabular}[c]{@{}l@{}}62.1\\ ±1.0\end{tabular}} & \begin{tabular}[c]{@{}l@{}}38.2\\ ±8.9\end{tabular} & \begin{tabular}[c]{@{}l@{}}51.4\\ ±2.2\end{tabular} & \textbf{\begin{tabular}[c]{@{}l@{}}63.5\\ ±5.1\end{tabular}} & \begin{tabular}[c]{@{}l@{}}43.1\\ ±2.3\end{tabular} & \textbf{\begin{tabular}[c]{@{}l@{}}73.3\\ ±5.7\end{tabular}} & \begin{tabular}[c]{@{}l@{}}63.9\\ ±0.3\end{tabular} & \begin{tabular}[c]{@{}l@{}}59.1\\ ±0.4\end{tabular} & \begin{tabular}[c]{@{}l@{}}83.1\\ ±0.6\end{tabular} & \begin{tabular}[c]{@{}l@{}}62.2\\ ±0.8\end{tabular} & \begin{tabular}[c]{@{}l@{}}16.2\\ ±2.7\end{tabular} & \textbf{\begin{tabular}[c]{@{}l@{}}91.8\\ ±0.3\end{tabular}} & \begin{tabular}[c]{@{}l@{}}47.4\\ ±6.6\end{tabular} & \multicolumn{1}{l|}{\textbf{\begin{tabular}[c]{@{}l@{}}47.0\\ ±0.7\end{tabular}}} & \begin{tabular}[c]{@{}l@{}}58.4\\ ±1.2\end{tabular} \\ \hline

        \multicolumn{21}{c}{Cityscapes$\rightarrow$ACDC (Test)}                                                                                                                                                                                                                                                                                                                                                  \\ \hline

        \multicolumn{1}{l|}{DAFormer \cite{daformer}}    & 58.4                                                & 51.3                                                & 84.0                                                & 42.7                                                & 35.1                                                & 50.7                                                & 30.0                                                & 57.0                                                & 74.8                                                & 52.8                                                & 51.3                                                & 58.3                                                & 32.6                                                & 82.7                                                & 58.3                                                & 54.9                                                & 82.4                                                & 44.1                                                & \multicolumn{1}{l|}{50.7}                                                & 55.4                                                \\
        \multicolumn{1}{l|}{HRDA \cite{hrda}}     & 88.3                                                & 57.9                                                & 88.1                                                & 55.2                                                & 36.7                                                & 56.3                                                & \textbf{62.9 }                                               & 65.3                                                & 74.2                                                & 57.7                                                & 85.9                                                & 68.8                                                & 45.7                                                & 88.5                                                & \textbf{76.4}                                                & 82.4                                                & 87.7                                                & 52.7                                                & \multicolumn{1}{l|}{60.4}                                                & 68.0                                                \\
        
        \multicolumn{1}{l|}{MIC \cite{mic}}       & \textbf{90.8}                                                & \textbf{67.1}                                                & \textbf{89.2}                                                & 54.5                                                & 40.5                                                & \textbf{57.2}                                                & 62.0                                                & \textbf{68.4 }                                               & 76.3                                                & 61.8                                                &\textbf{87.0}                                                & \textbf{71.3}                                                & \textbf{49.4}                                                &\textbf{89.7 }                                               & 75.7                                                & \textbf{86.8}                                                & \textbf{89.1}                                                & \textbf{56.9 }                                               & \multicolumn{1}{l|}{\textbf{63.0}}                                                & \textbf{70.4}                                                \\
        
        \multicolumn{1}{l|}{MIC+ECAP} & \begin{tabular}[c]{@{}l@{}}69.5\\ ±9.4\end{tabular} & \begin{tabular}[c]{@{}l@{}}56.9\\ ±5.1\end{tabular} & \textbf{\begin{tabular}[c]{@{}l@{}}89.2\\ ±0.2\end{tabular}} & \textbf{\begin{tabular}[c]{@{}l@{}}57.5\\ ±1.0\end{tabular}} & \textbf{\begin{tabular}[c]{@{}l@{}}43.8\\ ±1.4\end{tabular}} & \begin{tabular}[c]{@{}l@{}}56.4\\ ±0.4\end{tabular} & \begin{tabular}[c]{@{}l@{}}49.6\\ ±13.8\end{tabular} & \begin{tabular}[c]{@{}l@{}}67.2\\ ±0.6\end{tabular} & \textbf{\begin{tabular}[c]{@{}l@{}}77.0\\ ±1.0\end{tabular}} & \textbf{\begin{tabular}[c]{@{}l@{}}62.8\\ ±1.1\end{tabular}} & \begin{tabular}[c]{@{}l@{}}66.6\\ ±12.1\end{tabular} & \textbf{\begin{tabular}[c]{@{}l@{}}71.3\\ ±0.4\end{tabular}} & \begin{tabular}[c]{@{}l@{}}43.2\\ ±0.4\end{tabular} & \begin{tabular}[c]{@{}l@{}}89.4\\ ±0.4\end{tabular} & \begin{tabular}[c]{@{}l@{}}73.2\\ ±1.7\end{tabular} & \begin{tabular}[c]{@{}l@{}}68.0\\ ±3.8\end{tabular} & \begin{tabular}[c]{@{}l@{}}88.7\\ ±0.5\end{tabular} & \begin{tabular}[c]{@{}l@{}}56.8\\ ±0.6\end{tabular} & \multicolumn{1}{l|}{\begin{tabular}[c]{@{}l@{}}62.3\\ ±1.0\end{tabular}} & \begin{tabular}[c]{@{}l@{}}65.8\\ ±2.1\end{tabular} \\ \hline  
        
        \end{tabular}

        }
        \end{table*}


        We compare ECAP with existing UDA methods on four popular benchmarks in Table \ref{result:tab1}. 
        The reported mean and standard deviation of each experiment are computed from three runs with different random seeds. We use the following hyperparameters of ECAP: $n_0=1.0$, $\beta=0.95$, $n_c^{B_c} =40$ for GTA$\rightarrow$Cityscapes, $n_0=1.0$, $\beta=0.80$, $n_c^{B_c} =30$ for Synthia$\rightarrow$Cityscapes, $n_0=0.53$, $\beta=0.98$, $n_c^{B_c} =30$ for Cityscapes$\rightarrow$DarkZurich, and $n_0=1.0$, $\beta=0.90$, $n_c^{B_c} =50$ for Cityscapes$\rightarrow$ACDC.

        On GTA$\rightarrow$Cityscapes, ECAP gives a modest boost of $0.3$ mIoU to MIC. 
        On Synthia$\rightarrow$Cityscapes, we report the results of MIC as well as MIC$\dagger$, which is a variant of MIC that trains on pseudo-labels in the entire target image. 
        We find that MIC$\dagger$ outperforms MIC by 0.9 mIoU, showing that it is beneficial to train on the entire image for this benchmark. When additionally applying ECAP, performance is boosted further by another 0.9 mIoU, reaching an unprecedented performance of $69.1$ mIoU.

        On the other hand, ECAP degrades the performance of MIC on Cityscapes$\rightarrow$DarkZurich and Cityscapes$\rightarrow$ACDC due to large drops in IoU for certain classes, such as road and sidewalk, although some classes benefit from ECAP, especially wall and fence in Cityscapes$\rightarrow$ACDC. We hypothesize that ECAP is not as suitable for these domain adaptation benchmarks since the appearance of objects is a less discriminative factor for images with poor visibility. Instead, the context and prior knowledge about the scene becomes increasingly important to correctly segment images under such conditions. Since ECAP is based on aggressive data augmentation, it may hamper learning of context (e.g., the sky typically appears above the road in the images), which may be a significant issue in low visibility conditions. Qualitative results and an extended analysis is provided in the supplement.

    \subsection{ECAP on Other Methods}
    \label{sec:ecap_on_other_methods}
        We also implement ECAP on various prior art models on the GTA$\rightarrow$Cityscapes benchmark to understand how well ECAP generalizes across different methods. The results for each model without ECAP is taken from \cite{mic}, while the results with ECAP are computed from three experiments on different random seeds. In Table \ref{result:many_methods}, it can be seen that ECAP gives a substantial performance boost to all investigated methods, showing that it's not designed specifically for MIC. Noteably, both convolutional neural networks and transformer architectures benefit from ECAP. Furthermore, it is apparent that less capable models benefit more from ECAP than current state-of-the-art. This is expected since improving state-of-the-art becomes increasingly difficult as performance saturates. 
        \begin{table}[h!]
        \caption{Performance of different UDA methods without and with ECAP (mIoU in \%).}
        \label{result:many_methods}
        \resizebox{0.9\linewidth}{!}{
        \begin{tabular}{lllll}
        \hline
        Network   & UDA Method & w/o ECAP & w/ ECAP & diff \\ \hline
        DeepLabV2 & DACS       & 53.9     & 58.3    & +4.4 \\
        DeepLabV2 & DAFormer   & 56.0     & 61.2    & +5.2 \\
        DeepLabV2 & HRDA       & 63.0     & 65.6    & +2.6 \\
        DeepLabV2 & MIC        & 64.2     & 66.3    & +2.1 \\ \hline
        DAFormer  & DAFormer   & 68.3     & 69.1    & +0.8 \\
        DAFormer  & HRDA       & 73.8     & 75.0    & +1.2 \\
        DAFormer  & MIC        & 75.9     & 76.2    & +0.3 \\ \hline
        \end{tabular}
        }
        \end{table}

    \subsection{In-Depth Analysis of ECAP}
    \label{exp:ecap_indepth}
        This section provides an in-depth analysis of ECAP and specifically addresses the issue of pseudo-label noise, which has been the main driver of our proposed method. To save time, experiments are done on a single random seed and conducted with DAFormer since it has a substantially faster training time than MIC. In Table \ref{tab:pseudo-label-noise}, the results from training four different variants of DAFormer on GTA$\rightarrow$Cityscapes are presented.
        Two of the columns correspond to standard DAFormer and DAFormer+ECAP (with the same settings as in Section \ref{sec:ecap_on_other_methods}). On the other hand, DAFormer (denoise) uses the target domain labels to set the pixel-level weights to zero for any incorrect pseudo-labels, thereby eliminating such pseudo-labels' contribution to the training (recall that a weighted cross-entropy loss function is used). 
        Conversely, DAFormer (oracle) directly replaces all pseudo-labels with the corresponding labels, serving as an upper bound for DAFormer's performance. Along with mIoU of the four models, we also present the \textit{target accuracy}, which is the proportion of correct pseudo-labels, as well as the \textit{target loss noise ratio}, which is the proportion of the target loss that is derived from incorrect pseudo-labels. The target loss refers to the part of the loss that is derived from target domain pixels, which may originate from either the sampled target image or the ECAP memory bank. The target accuracy and target loss noise ratio are computed on the mixed training images and are averaged over the final 50 iterations of training. 
        
        In Table \ref{tab:pseudo-label-noise}, the performance of DAFormer (oracle) is 5.8 mIoU points higher than DAFormer. Moreover, DAFormer (denoise) significantly narrows this gap by 4.0 mIoU points, demonstrating that shifting focus towards correct pseudo-labels during training is an effective strategy. Furthermore, ECAP increases the target accuracy of DAFormer to a level on par with DAFormer (denoise) and additionally lowers the target loss noise ratio of DAFormer. This indicates that augmenting the training examples with ECAP increases the proportion of correctly pseudo-labeled content and as a result, the remaining erroneous pseudo-labels make up a smaller proportion of the loss value. Importantly, ECAP increases the mIoU of DAFormer significantly, although not as much as DAFormer (denoise) and (oracle) which both have access to the target domain labels.

        \begin{table}[h!]
        \caption{Performance (in \%) of four variants of DAFormer.}
        \label{tab:pseudo-label-noise}

        \resizebox{\linewidth}{!}{
        \begin{tabular}{lllll}
        \hline
         & DAFormer & \begin{tabular}[c]{@{}l@{}}DAFormer\\ (denoise)\end{tabular} &  \begin{tabular}[c]{@{}l@{}}DAFormer\\ (oracle)\end{tabular} &  \begin{tabular}[c]{@{}l@{}}ECAP\\ (DAFormer) \end{tabular} \\ \hline
         mIoU & 68.0 & 72.0           & 73.8 & 68.7 \\  
        \begin{tabular}[c]{@{}l@{}}Target accuracy\end{tabular} & 87.8 & 89.3           & 100.0 & 89.6 \\  
         \begin{tabular}[c]{@{}l@{}}Target loss noise ratio\end{tabular} & 37.9 & 0.0           & 0.0 & 33.3\\  \hline     
        \end{tabular}}
        \end{table}

        Table \ref{tab:2} further shows the target accuracy of DAFormer+ECAP (from Table \ref{tab:pseudo-label-noise}) split over different classes and pixels originating from the ECAP memory bank and the sampled target image respectively. It is evident that the content from the memory bank typically is associated with more accurate pseudo-labels. This is especially true for \textit{thing classes} such as train, motorbike and bike, while not as apparent for \textit{stuff classes} such as road, sidewalk and building. 
        
        
        \begin{table}[h!]
        \caption{Accuracy (in \%) of pseudo-labels for pixels originating from the ECAP memory bank and the sampled target image.}
        \label{tab:2}
        \resizebox{\linewidth}{!}{
        \begin{tabular}{lllllllll}
        \hline
                     & Road & Sidew. & Build. & ... & Train & M.bike & Bike \\ \hline 
        Memory bank  & 94.3 & 67.3   &  89.9   &     & 96.1  & 80.7   &  76.3 \\ 
        Target image & 91.9 & 60.5   & 90.4   &     & 55.0  & 37.5   & 57.4 \\ \hline 
        \end{tabular}}
        \end{table}

    \subsection{Hyperparameter Sensitivity Analysis}
    \label{exp:ecap_analysis}

            This section provides a sensitivity analysis of the hyperparameters of ECAP. To save time, we perform experiments on a single seed on GTA$\rightarrow$Cityscapes using DAFormer+ECAP ($n_0=1.0, \beta=0.93, n_c^{B_c} =30$) as a baseline and change the hyperparameters one at a time. 
            Table \ref{result:ablation} reports the performance and deviation from this baseline in terms of mIoU under a number of different settings. 

            \begin{table}[ht]
            \caption{Performance of ECAP (mIoU \%) under different hyperparameter settings. The $\Delta$-row displays the deviation from the ECAP baseline and ECAP$^-$ indicates the removal of random scaling, translation and flipping in the augmentation module.}
            \label{result:ablation}
            
            \resizebox{\linewidth}{!}{
            
            \begin{tabular}{l|ll|ll|lll|lll}
            \hline
            \vspace{-10pt}
             & \multicolumn{2}{c}{$n_0$} & \multicolumn{2}{c}{$\beta$} & \multicolumn{3}{c}{$n_c^{B_c}$} & &  &      \\

                     & 0.053        & 0.53       & 0.75           & 0.97       & 5        & 50       & 100   &   \begin{sideways} ECAP$^-$ \end{sideways}        &   \begin{sideways} DAFormer \end{sideways}     &    \begin{sideways} ECAP \end{sideways}      \\ \hline
            $\Delta$ & -0.6         & +0.2       & $\pm$0.0       & -0.8       & -1.0     & +0.2     & -0.7  &   $\pm$0.0 & -1.3     & $\pm$0.0 \\ \hline
            mIoU     & 68.5         & 69.3       & 69.1           & 68.3       & 68.1     & 69.3     & 68.4  &  69.1      & 67.8     & 69.1     \\ \hline
            \end{tabular}}
            \end{table}


            \noindent
            \textbf{ECAP intensity $n_0$:}
                Since ECAP is reduced to DAFormer when $n_0=0$ it is expected that the performance approaches that of DAFormer as the value of $n_0$ decreases. In Table \ref{result:ablation}, it can be seen that $n_0=0.053$ achieves lower performance than the ECAP baseline, although still superior to DAFormer. However, $n_0=0.53$ performs slightly better than the ECAP baseline, which indicates that $n_0=1.0$ (which out of convenience was used in most experiments of our paper) may not be optimal. 

            \noindent
            \textbf{Sampling schedule $\beta$:}
                We analyze the impact of changing the value of $\beta$ which determines when ECAP sampling comes online during training. In practice $\beta=0.75$, $\beta=0.9$ (baseline) and $\beta=0.97$ implies that ECAP comes online roughly at iteration 3k, 8k and 20k respectively. 
                In Table \ref{result:ablation}, it can be seen that ECAP performs well even when starting sampling very early in training, indicating that doing ECAP augmentation with initially less confident pseudo-labels doesn't impede learning. On the other hand, an excessively large value of $\beta$ results in performance similar to DAFormer, which is expected since ECAP is reduced to DAFormer as $\beta$ approaches $1.0$.

            \noindent
            \textbf{Effective memory bank size $n_c^{B_c}$:}
            In Table \ref{result:ablation} it can be seen that $n_c^{B_c}=50$ performs slightly better than the baseline $n_c^{B_c}=30$, while $n_c^{B_c}=5$ and $n_c^{B_c}=100$ perform significantly worse than the baseline, although still advantageous in comparison to DAFormer. We hypothesize that using an unnecessarily small memory bank is suboptimal since it implies less diversity of the ECAP samples. 
            Conversely, an excessively large memory bank implies that less confident samples are included in the memory bank, which may lower ECAP's effectiveness in reducing pseudo-label noise.
            

            \noindent
            \textbf{Transformations in the augmentation module:}
                In Table \ref{result:ablation}, ECAP$^-$ denotes the removal of the random scaling, translation, and horizontal flipping included in the augmentation module of ECAP. We note that removing these components doesn't effect the performance in this experiment and is not essential to the functioning of ECAP.

\section{Conclusions}
In this paper, we presented ECAP, a data augmentation method designed to reduce the adverse effect of erroneous pseudo-labels for unsupervised domain adaptive semantic segmentation. 
By cut-and-pasting confident pseudo-labeled target samples from a memory bank, ECAP benefits training by shifting focus away from erroneous pseudo-labels.
Through comprehensive experiments, we demonstrate the effectiveness of our approach on synthetic-to-real domain adaptation. 
Notably, we boost the performance of the recent method MIC with $0.3$ mIoU on GTA$\rightarrow$Cityscapes and $1.8$ mIoU on Synthia$\rightarrow$Cityscapes, setting new state-of-the-art performance in both cases. 
Our experiments on day-to-nighttime and clear-to-adverse-weather domain adaptation benchmarks additionally highlights a limitation of ECAP. Namely that ECAP may hamper the learning of context information and generate predictions with less bias and higher variance following training with aggressive data augmentation. Therefore, we find ECAP less suitable for adaptation to domains with e.g., poor visibility, where context information and a strong bias is pivotal for making accurate predictions.
Thanks to the demonstrated benefits of ECAP on synthetic-to-real UDA, we hope that ECAP can be part of future UDA methods to further push the state-of-the-art on this important problem. 

\medskip
\noindent
\textbf{Acknowledgment.} This work was supported by AB Volvo and the Wallenberg AI, Autonomous Systems and Software Program (WASP) funded by the Knut and Alice Wallenberg Foundation.
The experiments were enabled by resources provided by the National Academic Infrastructure for Supercomputing in Sweden (NAISS) and the Swedish National Infrastructure for Computing (SNIC) at Chalmers Centre for Computational Science and Engineering (C3SE) partially funded by the Swedish Research Council through grant agreements no. 2022-06725 and no. 2018-05973. 

\bibliographystyle{IEEEbib}
\bibliography{main}

\begin{thebibliography}{10}
\providecommand{\url}[1]{#1}
\csname url@samestyle\endcsname
\providecommand{\newblock}{\relax}
\providecommand{\bibinfo}[2]{#2}
\providecommand{\BIBentrySTDinterwordspacing}{\spaceskip=0pt\relax}
\providecommand{\BIBentryALTinterwordstretchfactor}{4}
\providecommand{\BIBentryALTinterwordspacing}{\spaceskip=\fontdimen2\font plus
\BIBentryALTinterwordstretchfactor\fontdimen3\font minus \fontdimen4\font\relax}
\providecommand{\BIBforeignlanguage}[2]{{%
\expandafter\ifx\csname l@#1\endcsname\relax
\typeout{** WARNING: IEEEtran.bst: No hyphenation pattern has been}%
\typeout{** loaded for the language `#1'. Using the pattern for}%
\typeout{** the default language instead.}%
\else
\language=\csname l@#1\endcsname
\fi
#2}}
\providecommand{\BIBdecl}{\relax}
\BIBdecl

\bibitem{cycada}
J.~Hoffman, E.~Tzeng, T.~Park, J.-Y. Zhu, P.~Isola, K.~Saenko, A.~Efros, and T.~Darrell, ``{C}y{CADA}: Cycle-consistent adversarial domain adaptation,'' in \emph{ICML}, vol.~80, 10--15 Jul 2018, pp. 1989--1998.

\bibitem{adaptsegnet}
Y.-H. Tsai, W.-C. Hung, S.~Schulter, K.~Sohn, M.-H. Yang, and M.~Chandraker, ``Learning to adapt structured output space for semantic segmentation,'' in \emph{CVPR}, June 2018.

\bibitem{advent}
T.-H. Vu, H.~Jain, M.~Bucher, M.~Cord, and P.~Perez, ``{ADVENT}: Adversarial entropy minimization for domain adaptation in semantic segmentation,'' in \emph{CVPR}, June 2019.

\bibitem{discriminative_path_repr}
Y.-H. Tsai, K.~Sohn, S.~Schulter, and M.~Chandraker, ``Domain adaptation for structured output via discriminative patch representations,'' in \emph{ICCV}, October 2019.

\bibitem{CBST}
Y.~Zou, Z.~Yu, B.~V. Kumar, and J.~Wang, ``Unsupervised domain adaptation for semantic segmentation via class-balanced self-training,'' in \emph{ECCV}, September 2018.

\bibitem{dacs}
W.~Tranheden, V.~Olsson, J.~Pinto, and L.~Svensson, ``{DACS}: Domain adaptation via cross-domain mixed sampling,'' in \emph{WACV}, January 2021, pp. 1379--1389.

\bibitem{iast}
K.~Mei, C.~Zhu, J.~Zou, and S.~Zhang, ``Instance adaptive self-training for unsupervised domain adaptation,'' in \emph{ECCV}.\hskip 1em plus 0.5em minus 0.4em\relax Springer, 2020, pp. 415--430.

\bibitem{proda}
P.~Zhang, B.~Zhang, T.~Zhang, D.~Chen, Y.~Wang, and F.~Wen, ``Prototypical pseudo label denoising and target structure learning for domain adaptive semantic segmentation,'' in \emph{CVPR}, June 2021, pp. 12\,414--12\,424.

\bibitem{hrda}
L.~Hoyer, D.~Dai, and L.~Van~Gool, ``{HRDA}: Context-aware high-resolution domain-adaptive semantic segmentation,'' in \emph{ECCV}, 2022, pp. 372--391.

\bibitem{mic}
L.~Hoyer, D.~Dai, H.~Wang, and L.~Van~Gool, ``{MIC}: Masked image consistency for context-enhanced domain adaptation,'' in \emph{CVPR}, June 2023, pp. 11\,721--11\,732.

\bibitem{daformer}
L.~Hoyer, D.~Dai, and L.~Van~Gool, ``{DAF}ormer: Improving network architectures and training strategies for domain-adaptive semantic segmentation,'' in \emph{CVPR}, 2022, pp. 9924--9935.

\bibitem{chen2022pipa}
M.~Chen, Z.~Zheng, Y.~Yang, and T.-S. Chua, ``Pipa: Pixel- and patch-wise self-supervised learning for domain adaptative semantic segmentation,'' in \emph{Proceedings of the 31st ACM International Conference on Multimedia}, 2023, p. 1905–1914.

\bibitem{crst}
Y.~Zou, Z.~Yu, X.~Liu, B.~V. Kumar, and J.~Wang, ``Confidence regularized self-training,'' in \emph{ICCV}, October 2019.

\bibitem{pycda}
Q.~Lian, F.~Lv, L.~Duan, and B.~Gong, ``Constructing self-motivated pyramid curriculums for cross-domain semantic segmentation: A non-adversarial approach,'' in \emph{ICCV}, October 2019.

\bibitem{sac}
N.~Araslanov and S.~Roth, ``Self-supervised augmentation consistency for adapting semantic segmentation,'' in \emph{CVPR}, June 2021, pp. 15\,384--15\,394.

\bibitem{dsp}
L.~Gao, J.~Zhang, L.~Zhang, and D.~Tao, ``{DSP}: Dual soft-paste for unsupervised domain adaptive semantic segmentation,'' in \emph{Proceedings of the 29th {ACM} International Conference on Multimedia}.\hskip 1em plus 0.5em minus 0.4em\relax {ACM}, oct 2021.

\bibitem{li2019bidirectional}
Y.~Li, L.~Yuan, and N.~Vasconcelos, ``Bidirectional learning for domain adaptation of semantic segmentation,'' in \emph{CVPR}, June 2019.

\bibitem{saito2018maximum}
K.~Saito, K.~Watanabe, Y.~Ushiku, and T.~Harada, ``Maximum classifier discrepancy for unsupervised domain adaptation,'' in \emph{CVPR}, June 2018.

\bibitem{cda}
Y.~Zhang, P.~David, and B.~Gong, ``Curriculum domain adaptation for semantic segmentation of urban scenes,'' in \emph{ICCV}.\hskip 1em plus 0.5em minus 0.4em\relax {IEEE}, oct 2017.

\bibitem{chen2019domain}
M.~Chen, H.~Xue, and D.~Cai, ``Domain adaptation for semantic segmentation with maximum squares loss,'' in \emph{ICCV}, 2019, pp. 2090--2099.

\bibitem{hiast}
C.~Zhu, K.~Liu, W.~Tang, K.~Mei, J.~Zou, and T.~Huang, ``Hard-aware instance adaptive self-training for unsupervised cross-domain semantic segmentation,'' \emph{arXiv preprint arXiv:2302.06992}, 2023.

\bibitem{huang2022category}
J.~Huang, D.~Guan, A.~Xiao, S.~Lu, and L.~Shao, ``Category contrast for unsupervised domain adaptation in visual tasks,'' in \emph{CVPR}, June 2022, pp. 1203--1214.

\bibitem{sohn2020fixmatch}
K.~Sohn, D.~Berthelot, N.~Carlini, Z.~Zhang, H.~Zhang, C.~A. Raffel, E.~D. Cubuk, A.~Kurakin, and C.-L. Li, ``{F}ix{M}atch: Simplifying semi-supervised learning with consistency and confidence,'' in \emph{NeurIPS}, vol.~33, 2020, pp. 596--608.

\bibitem{context-aware-mixup}
Q.~Zhou, Z.~Feng, Q.~Gu, J.~Pang, G.~Cheng, X.~Lu, J.~Shi, and L.~Ma, ``Context-aware mixup for domain adaptive semantic segmentation,'' \emph{IEEE Transactions on Circuits and Systems for Video Technology}, vol.~33, no.~2, pp. 804--817, 2023.

\bibitem{cag}
Q.~ZHANG, J.~Zhang, W.~Liu, and D.~Tao, ``Category anchor-guided unsupervised domain adaptation for semantic segmentation,'' in \emph{NeurIPS}, vol.~32, 2019.

\bibitem{uncertainty_aware_consistency_regularization}
Q.~Zhou, Z.~Feng, Q.~Gu, G.~Cheng, X.~Lu, J.~Shi, and L.~Ma, ``Uncertainty-aware consistency regularization for cross-domain semantic segmentation,'' \emph{Computer Vision and Image Understanding}, vol. 221, p. 103448, 2022.

\bibitem{rectifying_with_uncertainty}
Z.~Zheng and Y.~Yang, ``Rectifying pseudo label learning via uncertainty estimation for domain adaptive semantic segmentation,'' \emph{IJCV}, vol. 129, no.~4, pp. 1106--1120, 2021.

\bibitem{corda}
Q.~Wang, D.~Dai, L.~Hoyer, L.~Van~Gool, and O.~Fink, ``Domain adaptive semantic segmentation with self-supervised depth estimation,'' in \emph{ICCV}, October 2021, pp. 8515--8525.

\bibitem{cutmix}
S.~Yun, D.~Han, S.~J. Oh, S.~Chun, J.~Choe, and Y.~Yoo, ``{C}ut{M}ix: Regularization strategy to train strong classifiers with localizable features,'' in \emph{ICCV}, October 2019.

\bibitem{cut_paste_learn}
D.~Dwibedi, I.~Misra, and M.~Hebert, ``Cut, paste and learn: Surprisingly easy synthesis for instance detection,'' in \emph{ICCV}, 2017, pp. 1301--1310.

\bibitem{modeling_context_key_to_OD}
N.~Dvornik, J.~Mairal, and C.~Schmid, ``Modeling visual context is key to augmenting object detection datasets,'' in \emph{ECCV}, 2018, pp. 364--380.

\bibitem{ge2022empaste}
Y.~Ge, J.~Xu, B.~N. Zhao, L.~Itti, and V.~Vineet, ``{EM-Paste}: {EM}-guided cut-paste with {DALL-E} augmentation for image-level weakly supervised instance segmentation,'' \emph{arXiv preprint arXiv:2212.07629}, 2022.

\bibitem{classmix}
V.~Olsson, W.~Tranheden, J.~Pinto, and L.~Svensson, ``{C}lass{M}ix: Segmentation-based data augmentation for semi-supervised learning,'' in \emph{WACV}, January 2021, pp. 1369--1378.

\bibitem{simple_copy_paste_instance_segmentation}
G.~Ghiasi, Y.~Cui, A.~Srinivas, R.~Qian, T.-Y. Lin, E.~D. Cubuk, Q.~V. Le, and B.~Zoph, ``Simple copy-paste is a strong data augmentation method for instance segmentation,'' in \emph{CVPR}, 2021, pp. 2918--2928.

\bibitem{zhao2022xpaste}
H.~Zhao, D.~Sheng, J.~Bao, D.~Chen, D.~Chen, F.~Wen, L.~Yuan, C.~Liu, W.~Zhou, Q.~Chu, W.~Zhang, and N.~Yu, ``X-paste: Revisiting scalable copy-paste for instance segmentation using {CLIP} and {S}table{D}iffusion,'' in \emph{ICML}, vol. 202, 23--29 Jul 2023, pp. 42\,098--42\,109.

\bibitem{ida}
Z.~Chen, Z.~Ding, J.~M. Gregory, and L.~Liu, ``{IDA}: Informed domain adaptive semantic segmentation,'' \emph{arXiv preprint arXiv:2303.02741}, 2023.

\bibitem{Richter_2016_ECCV}
S.~R. Richter, V.~Vineet, S.~Roth, and V.~Koltun, ``Playing for data: {G}round truth from computer games,'' in \emph{ECCV}, vol. 9906, 2016, pp. 102--118.

\bibitem{Ros_2016_CVPR}
G.~Ros, L.~Sellart, J.~Materzynska, D.~Vazquez, and A.~M. Lopez, ``The {SYNTHIA} dataset: A large collection of synthetic images for semantic segmentation of urban scenes,'' in \emph{CVPR}, June 2016.

\bibitem{cityscapes}
M.~Cordts, M.~Omran, S.~Ramos, T.~Rehfeld, M.~Enzweiler, R.~Benenson, U.~Franke, S.~Roth, and B.~Schiele, ``The cityscapes dataset for semantic urban scene understanding,'' in \emph{CVPR}, 2016.

\bibitem{darkzurich}
C.~Sakaridis, D.~Dai, and L.~Van~Gool, ``Guided curriculum model adaptation and uncertainty-aware evaluation for semantic nighttime image segmentation,'' in \emph{ICCV}, 2019.

\bibitem{acdc}
------, ``{ACDC}: The adverse conditions dataset with correspondences for semantic driving scene understanding,'' in \emph{ICCV}, 2021, pp. 10\,765--10\,775.

\end{thebibliography}

\setcounter{section}{0}

\newpage
\renewcommand{\thesection}{\Alph{section}}
\section*{Supplementary}

In this supplementary material, we analyze the predictions of ECAP qualitatively, motivate the introduction of the variant of MIC denoted as MIC$\dagger$, and provide an extended analysis of our method and its limitations.

\section{Qualitative Comparison with State-of-the-Art}

    \begin{figure*}[]
        \centering
        \includegraphics[width=0.99\linewidth]{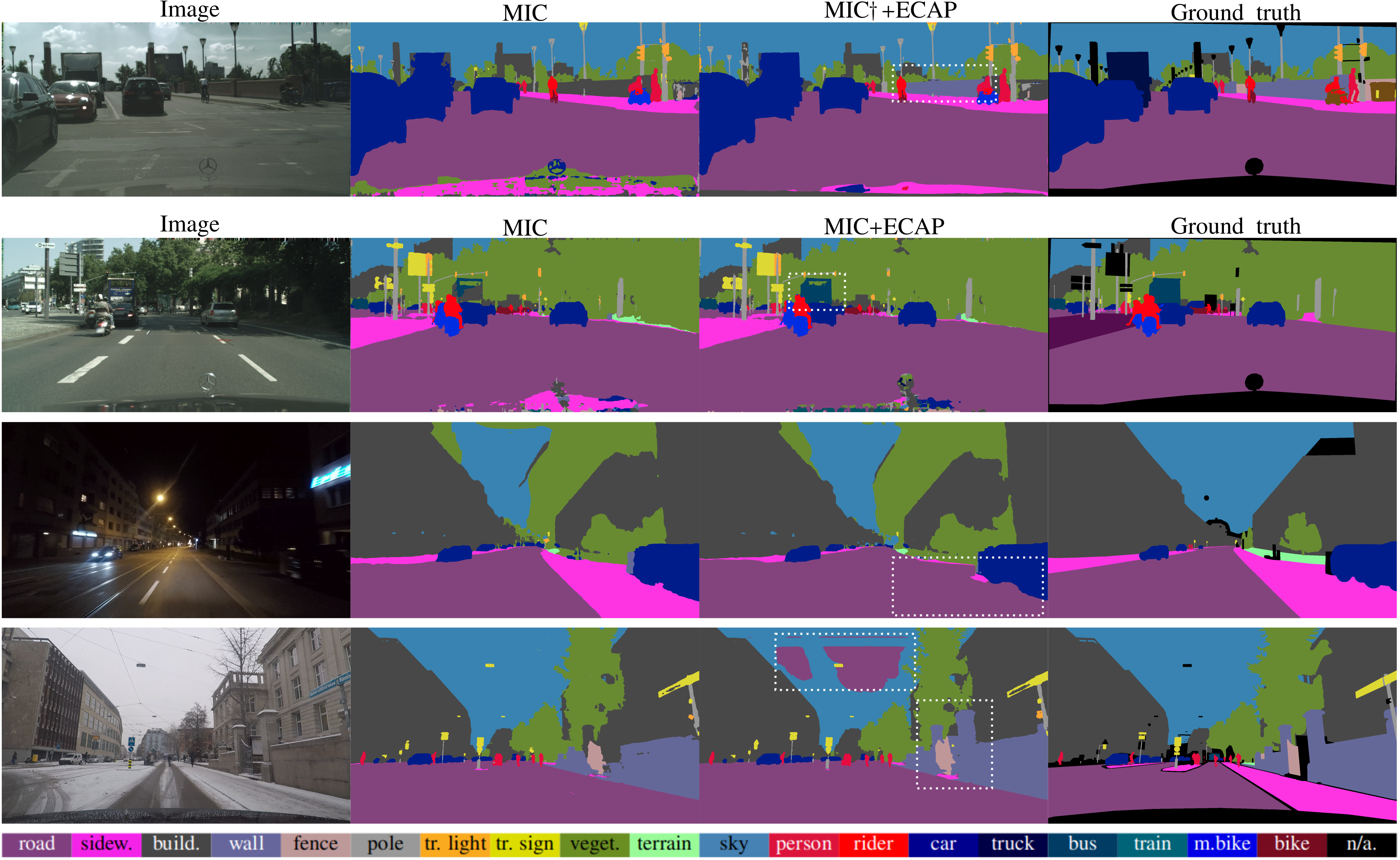}
        \caption{Qualitative comparison of MIC and MIC$\dagger$+ECAP on Synthia$\rightarrow$Cityscapes (row 1), as well as MIC and MIC+ECAP on GTA$\rightarrow$Cityscapes (row 2), Cityscapes$\rightarrow$DarkZurich (row 3), and Cityscapes$\rightarrow$ACDC (row 4).}
        \label{fig:qualitative}
    \end{figure*}
    
    In this section, we compare the predictions of MIC+ECAP with those of MIC qualitatively. Specifically, Figure \ref{fig:qualitative} shows the predictions of the two methods on one image from the (target domain) validation set for each of the evaluated benchmarks. Notably, in row 1 and 2, MIC+ECAP predicts more accurate masks for the classes wall and bus on Synthia$\rightarrow$Cityscapes and GTA$\rightarrow$Cityscapes respectively. Row 3 and 4 further illustrates failure cases of MIC+ECAP on Cityscapes$\rightarrow$DarkZurich and Cityscapes$\rightarrow$ACDC respectively. Specifically, in row 3, MIC+ECAP misclassifies the sidewalk as road, and in row 4, MIC+ECAP misclassifies the sky as road. These observations qualitatively explain the large drops in IoU for the classes \textit{sidewalk} on Cityscapes$\rightarrow$DarkZurich and \textit{sky} and \textit{road} on \newline
    Cityscapes$\rightarrow$ACDC that were presented in the main paper.

\section{Motivation of MIC$\dagger$}

    While DAFormer, HRDA and MIC refrain from training with pseudo-labels on the regions of the image corresponding to the ego-vehicle hood and the image boarders on GTA$\rightarrow$Cityscapes and Synthia$\rightarrow$Cityscapes, we find it beneficial to train with pseudo-labels in the entire image on Synthia$\rightarrow$Cityscapes. As shown in the main paper, letting MIC train on pseudo-labels in the entire image (denoted by MIC$\dagger$) increases the performance by 0.9 mIoU on Synthia$\rightarrow$Cityscapes. In this section we provide an analysis of this phenomenon.

    \begin{figure*}[]
        \centering
        \includegraphics[width=0.99\linewidth]{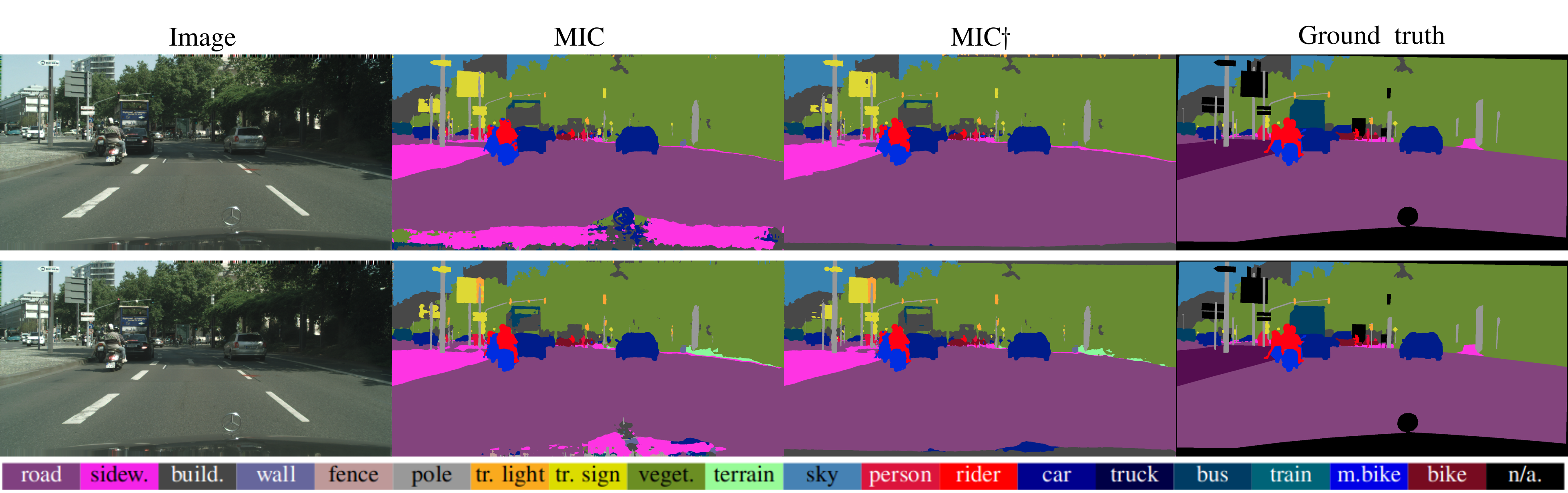}
        \caption{Predictions on Cityscapes validation images following training on Synthia$\rightarrow$Cityscapes (row 1) and GTA$\rightarrow$Cityscapes (row 2).}
        \label{fig:vehicle_hood}
    \end{figure*}

    The first row in Figure \ref{fig:vehicle_hood} shows the predictions of MIC and MIC$\dagger$ on Synthia$\rightarrow$Cityscapes, where MIC refrains from training on pseudo-labels in the mentioned regions (following the implementation of MIC) and MIC$\dagger$ instead trains on the entire target image. It can be seen that 
    MIC makes ambiguous predictions while MIC$\dagger$ typically predicts the class \textit{road} in the region corresponding to the ego-vehicle hood. Since this region is ignored during evaluation on the Cityscapes benchmark, this may seem like an insignificant detail. However, the ambiguous predictions of MIC are more prone to spilling over from the ego-vehicle hood to the road ahead. Furthermore, the ambiguous predictions in this region may correspond to rare classes such as bus, in which case these predictions may have a significant impact on the resulting mIoU score. By training on the pseudo-labels in this region, the predictions become more stable and tend not to spill over to the road ahead.

    Although training on the whole image is beneficial for\newline Synthia$\rightarrow$Cityscapes, it is not for GTA$\rightarrow$Cityscapes. The second row in \ref{fig:vehicle_hood} shows predictions of  MIC and MIC$\dagger$ on GTA$\rightarrow$Cityscapes. Also in this case, the predictions of MIC$\dagger$ are less sporadic than those of MIC in the region of the ego-vehicle hood. However, the predictions of MIC are not as prone to spilling over from the ego-vehicle hood when training on GTA$\rightarrow$Cityscapes as they are when training on Synthia$\rightarrow$Cityscapes. Therefore, it is not necessary to train on the pseudo-labels in this region for the GTA$\rightarrow$Cityscapes benchmark. In fact, MIC$\dagger$ achieves an average mIoU score of $75.26$ over three random seeds, which is inferior to the score of $75.9$ achieved by MIC on GTA$\rightarrow$Cityscapes. Apparently, it is better not to train on this region on GTA$\rightarrow$Cityscapes, which probably is the reason why MIC adopts this strategy.
    
    We hypothesize that training on pseudo-labels on the ego-vehicle hood is non-informative for the actual evaluation task, making it beneficial to ignore this region, unless this leads to unexpected problems as for Synthia$\rightarrow$Cityscapes. Furthermore, we believe that the predictions on the ego-vehicle hood are more prone to spilling over in the case of Synthia$\rightarrow$Cityscapes since there is typically no ego-vehicle hood in the Synthia dataset. Additionally, due to the perspective of the camera in Synthia, virtually any object class can appear in this region of the image. In the GTA dataset on the other hand, there is an ego-vehicle hood and typically a road segment around the ego-vehicle hood. This makes training on GTA more prone to predicting a sharp edge between ego-vehicle hood and the surrounding road segment, while training on Synthia results in more ambiguous predictions in this region. In the Dark Zurich and ACDC datasets, there exists no ego-vehicle hood in the test images, and the issue is avoided completely.

\section{ECAP Extended Analysis}
\begin{figure}[]
        \centering
        \includegraphics[width=0.99\linewidth]{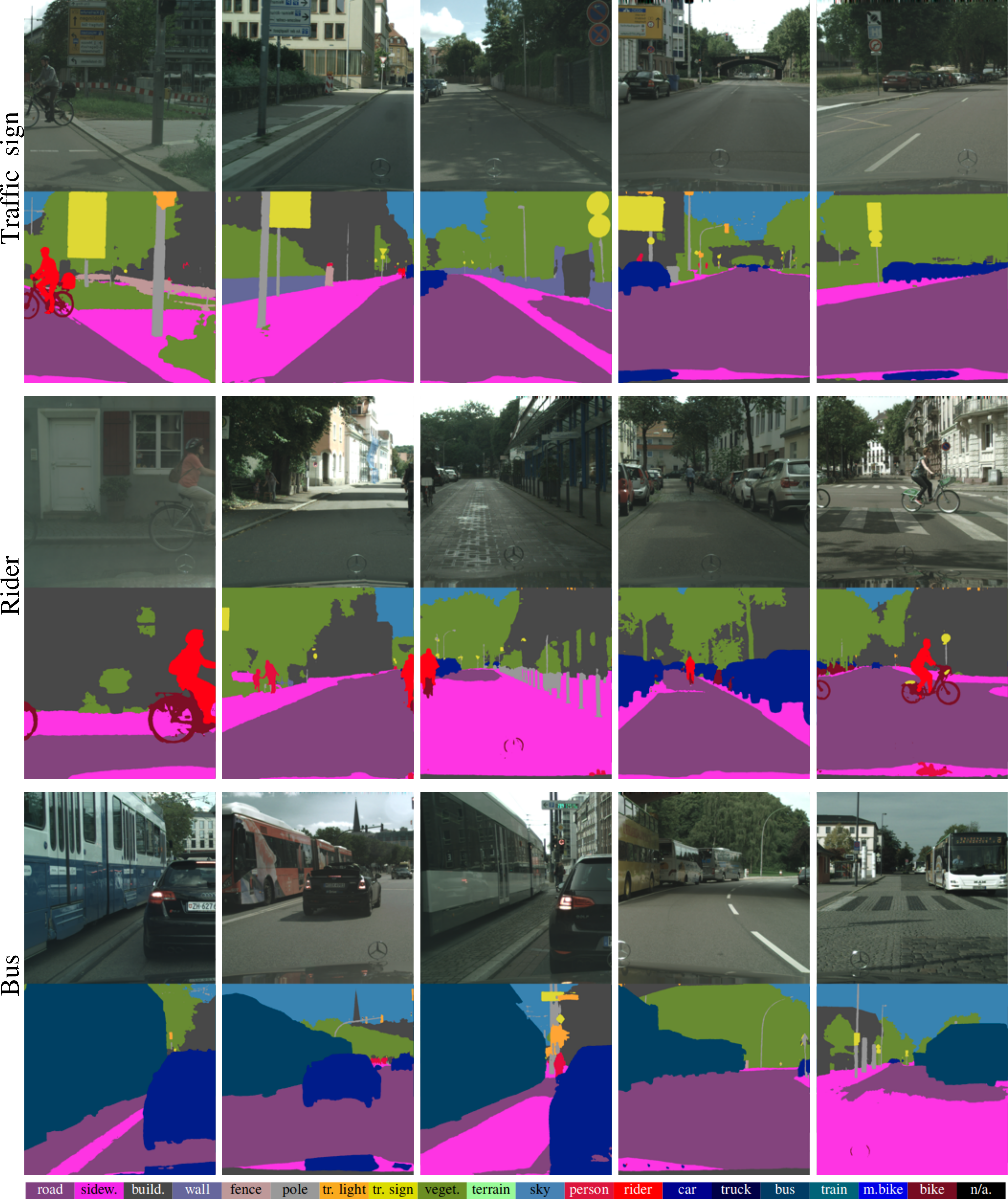}
        \caption{The five most confident samples of the classes \textit{traffic sign} (row 1), \textit{rider} (row 2), and \textit{bus} (row 3) in the ECAP memory bank. The samples are present in the memory bank at the end of training in the MIC+ECAP run on Synthia$\rightarrow$Cityscapes with median performance.}
        \label{fig:memory_bank}
    \end{figure}

    To gain a better understanding of ECAP, we study the memory bank in detail in this section. Figure \ref{fig:memory_bank} shows the five most confident samples, along with associated pseudo-labels, of each of the classes traffic sign, rider and bus in the memory bank of the median run of MIC$\dagger$+ECAP on Synthia$\rightarrow$Cityscapes presented in the main paper. It can be noted that the instances of the respective classes often are clearly visible in the images and are relatively close to the camera, which presumably is the reason why these samples generally are associated with high-quality pseudo-labels. It should also be noted that the images of the memory bank are not full-sized images. The reason for this is that they originate from the sampled target images in every iteration, which are cropped to size $1024\times 1024$. Additionally, since the class \textit{train} is not present in Synthia, it is reasonable that these are misclassified as \textit{bus} on Synthia$\rightarrow$Cityscapes, which explains why some trains are present in the third row of Figure \ref{fig:memory_bank}.

\begin{figure}[]
        \centering
        \includegraphics[width=0.5\linewidth]{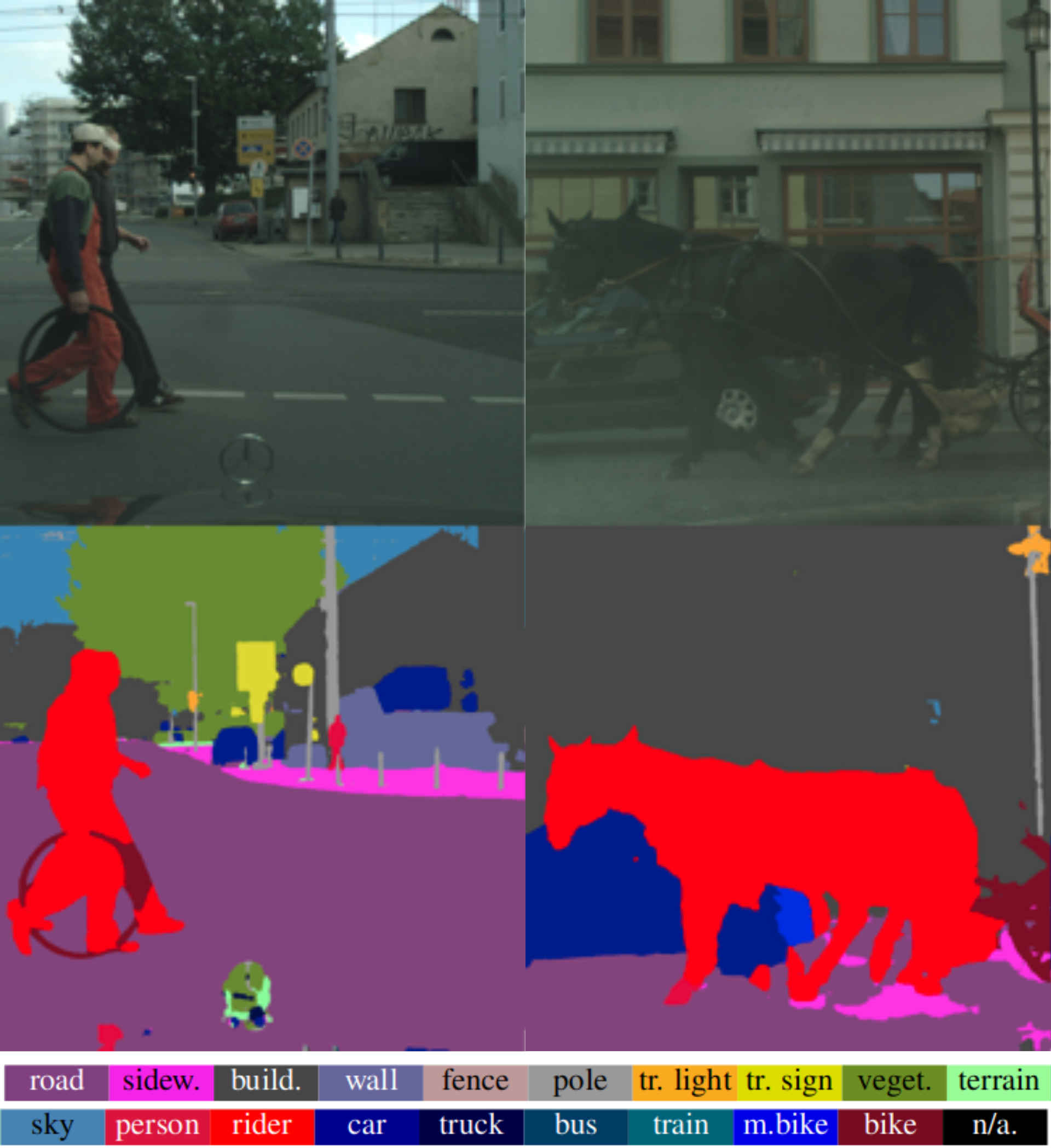}
        \caption{Two images (row 1) in the memory bank of class \textit{rider} that has been assigned inaccurate pseudo-labels (row 2) during the MIC+ECAP run on GTA$\rightarrow$Cityscapes with median performance.}
        \label{fig:bank_error}
    \end{figure}

    Although the memory bank generally provides high quality pseudo-labels, which is a result from the positive correlation between accuracy and confidence, it also inevitably include some erroneous pseudo-labels. In Figure \ref{fig:bank_error}, two samples from the memory bank of class \textit{rider} of the MIC+ECAP run on GTA with median performance are shown. These two samples where present in the memory bank in the end of the training and both display a high confidence of the class rider ($\approx 0.97)$ although both are misclassified examples. This highlights a problem with using predicted confidence to identify accurate predictions, namely that even misclassified examples may display a high confidence. Intuitively, this constitutes a potential risk of ECAP as erroneous pseudo-labels may be used excessively in ECAP augmentation.

\begin{figure}[]
        \centering
        \includegraphics[width=0.99\linewidth]{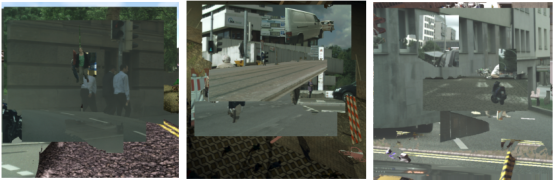}
        \caption{Augmented training examples of MIC$\dagger$+ECAP run on Synthia$\rightarrow$Cityscapes with median performance.}
        \label{fig:augmented_samples}
    \end{figure}

    Figure \ref{fig:augmented_samples} shows three training samples that have been generated through ECAP augmentation. The samples consist of a mix between the source and target images sampled in the current iteration as well as multiple samples from the memory bank. A few things are worth pointing out. First, many classes are present in the images since multiple classes are cut-and-pasted from the memory bank. This may facilitate learning different classes in the UDA setting and counteract the problem of self-training being dominated by pseudo-labels of certain easy-to-adapt classes. Second, when cut-and-pasting classes from the memory bank, they end up in a new context and the resulting images are typically unrealistic. While this could have the benefit of better learning to detect classes in unusual contexts, it may also be a drawback since it hampers learning of context information and certain prior knowledge.

\section{Limitations}
    The results of the main paper indicate that ECAP is not beneficial for day-to-nighttime (Cityscapes$\rightarrow$DarkZurich) or clear-to-adverse-weather (Cityscapes$\rightarrow$ACDC) unsupervised domain adaptation. As illustrated in Figure \ref{fig:qualitative}, ECAP struggles with, for example, sidewalk, road and sky, while MIC handles these classes better. This does not, however, apply to the benchmarks on synthetic-to-real domain adaptation. We believe that one reason for this is that the appearance of the road and sidewalk becomes more similar during nighttime. Similarly, a clouded sky has similar appearance as a snow-covered road, making it more difficult to distinguish them from each other than in clear weather conditions. Therefore, learning context information becomes increasingly important to attain good segmentation results on Cityscapes$\rightarrow$DarkZurich and Cityscapes$\rightarrow$ACDC. On the other hand, context information is of little importance in the augmented training examples of ECAP, which encourages the model to instead focus on the appearance of classes during training. In this sense, ECAP may hamper the learning of context information, and as a result, struggles with images in which context information and prior knowledge of the scene layout is pivotal for making an accurate segmentation. 
    


\end{document}